\documentclass[journal]{IEEEtran}

\usepackage{graphicx}
\usepackage{amsfonts}

\usepackage[english]{babel}
\usepackage{amsmath, amssymb}
\usepackage{float}
\usepackage{makecell}
\usepackage{multirow}
\usepackage{cite}
\usepackage{amsmath,amssymb,amsfonts}
\usepackage{algorithmic}
\usepackage{graphicx}
\usepackage{textcomp}
\usepackage{xcolor}

\usepackage{subfigure}
\usepackage{epstopdf}

\usepackage{graphicx}
\usepackage{comment}
\usepackage{amsmath,amssymb} 
\usepackage{color}
\usepackage{enumerate}

\usepackage{booktabs}

\ifCLASSINFOpdf
\else
\fi
\begin{document}

\title{
Interactive Multi-scale Fusion of 2D and 3D Features for Multi-object Tracking
}

\author{Guangming Wang, Chensheng Peng, Jinpeng Zhang, and Hesheng Wang
        
\thanks{*This work was supported in part by the Natural Science Foundation of China under Grant 62073222 and U1913204, in part by “Shu Guang” project supported by Shanghai Municipal Education Commission and Shanghai Education Development Foundation under Grant 19SG08, in part by Shenzhen Science and Technology Program under Grant JSGG20201103094400002, in part by the Science and Technology Commission of Shanghai Municipality under Grant 21511101900, in part by grants from NVIDIA Corporation. Corresponding Author: Hesheng Wang. The first two authors contributed equally.}
\thanks{G. Wang, C. Peng, and H. Wang are with Department of Automation, Key Laboratory of System Control and Information Processing of Ministry of Education, Key Laboratory of Marine Intelligent Equipment and System of Ministry of Education, Shanghai Engineering Research Center of Intelligent Control and Management, Shanghai Jiao Tong University, Shanghai 200240, China. (e-mail: wanghesheng@sjtu.edu.cn).

J. Zhang is with XLAB, the Second Academy of China Aerospace Science And Industry Corporation, Beijing 100854, China.

}

}

%
%

\markboth{Journal of \LaTeX\ Class Files,~Vol.~14, No.~8, August~2015}%
{Shell \MakeLowercase{\textit{et al.}}: Bare Demo of IEEEtran.cls for IEEE Journals}
%



\maketitle


\begin{abstract}
Multiple object tracking (MOT) is a significant task in achieving autonomous driving. Traditional works attempt to complete this task, either based on point clouds (PC) collected by LiDAR, or based on images captured from cameras. However, relying on one single sensor is not robust enough, because it might fail during the tracking process. On the other hand, feature fusion from multiple modalities contributes to the improvement of accuracy. As a result, new techniques based on different sensors integrating features from multiple modalities are being developed. Texture information from RGB cameras and 3D structure information from Lidar have respective advantages under different circumstances. However, it's not easy to achieve effective feature fusion because of completely distinct information modalities. Previous fusion methods usually fuse the top-level features after the backbones extract the features from different modalities. In this paper, we first introduce PointNet++ to obtain multi-scale deep representations of point cloud to make it adaptive to our proposed Interactive Feature Fusion between multi-scale features of images and point clouds. Specifically, through multi-scale interactive query and fusion between pixel-level and point-level features, our method, can obtain more distinguishing features to improve the performance of multiple object tracking. Besides, we explore the effectiveness of pre-training on each single modality and fine-tuning on the fusion-based model. The experimental results demonstrate that our method can achieve good performance on the KITTI benchmark and outperform other approaches without using multi-scale feature fusion. Moreover, the ablation studies indicates the effectiveness of multi-scale feature fusion and pre-training on single modality.

\end{abstract}

\IEEEpeerreviewmaketitle

\section{Introduction}

Multiple object tracking (MOT) is a crucial component of autonomous driving. Multi-object tracking is about keeping the identification (ID) propagation for the same object and forming new tracklets in  a video sequence. It is an essential part of many applications, such as self-driving, odometry and robot collision prediction \cite{xiang2020end,zhou2018deep}.

Autonomous vehicles employ a variety of sensors, including stereo cameras, LiDAR, Radar sensors, etc. Accordingly, various algorithms are developed to accomplish the MOT task based on the aforementioned sensors, primarily cameras and LiDAR\cite{BeyondPixels,bergmann2019tracking,2019MOTS,weng20203d}. The methods \cite{bergmann2019tracking,2019MOTS} in the image domain take the RGB images captured by the camera as inputs, then apply convolutional neural networks (CNNs), such as VGG-Net\cite{simonyan2014very} and ResNet\cite{ResNet} for feature extraction. Other studies\cite{weng20203d} are based on LiDAR, which have attracted more and more attention as a result of the development of deep representation of point cloud \cite{PointNet,PointNet++} and 3D object detection\cite{shi2019pointrcnn}. The development of object detection in 3D spaces promotes the feature extraction of point clouds, which also benefits the feature extraction process in the object tracking task.

However, all of the methods based on a single modality have several drawbacks in common, such as low precision and lack of reliability because occlusion and fast moving instances tend to cause mismatching problems. Therefore, a number of methods\cite{BeyondPixels,CenterTrack} based on different modalities are developed. A multi-modality MOT framework, mmMOT \cite{mmmot}, is proposed, where features from LiDAR and images are extracted separately and then fused together using a fusion module to form a single feature. In order to combine image and LiDAR features, GNN3DMOT\cite{GNN3DMOT} provides two operations:  1) Addition of 2D and 3D features, 2) Concatenation of 2D and 3D features.  EagerMOT \cite{EagerMOT} establishes a matching between 2D and 3D detections by projecting the 3D detections to the image plane to avoid false detection, but it does not pay much attention to the feature fusion.  All of these methods only fuse the top-level features obtained from the backbones once. However, the abundant information embedded in the low-level features and the position relationship in the original pixels and point clouds are ignored, which is significant for extracting as much as information from different sensors. 

To fully exploit the features from different levels, in the paper, we design an multi-scale fusion module of 2D and 3D features for multiple object tracking. Inspired by the idea of projection of 3D detections on the image plane from \cite{EagerMOT}, our method is able to achieve interactive feature fusion between image and LiDAR based on the positional relationship of the projected point clouds and pixels on the image plane. During the fusion process, the features of grouped points (or pixels) are fused with the center pixels (or points) using attention mechanism, where the features of every point (or pixel) are assigned a weight value to represent its importance. Apart from four times of multi-scale feature fusion, our method also further exploits the inherent position information between point clouds and corresponding image patches to gain better performance, which previous methods\cite{mmmot,GNN3DMOT} ignored in the fusion process. 

In recent years, a variety of studies have shown that pre-trained models (PTMs) on ImageNet\cite{deng2009imagenet} can learn generic image representations, which is helpful for downstream computer vision tasks. Pre-training can help avoid training models from scratch. After we design a network, which is usually a multi-layer convolutional neural networks (CNN) for images, we can use one training set to pre-train this network, learn the network parameters on one task, and then store them for further use. For another downstream task, the network parameters can be initialized by loading the parameters learned in the first task then fine-tuned on the new dataset. Different from traditional pre-training, in this paper, we first pre-train our models on single modality, images and point clouds respectively. After pre-training, we take the parameters of the two feature extraction backbones and use them for the model based on multiple modalities.

In summary, our main
contributions are as follows:

\begin{itemize} 
    \item A novel interactive feature fusion module is proposed to achieve interactive fusion between features from different modalities by exploiting the inherent positional relationship between point clouds and pixels of images. 
    \item The multi-scale feature fusion process is performed for four times in total during the feature extraction process. Both the low-level and high-level features are used for the final prediction.
    \item Effectiveness of pre-training in multi-object tracking task is explored. We first train the model based on one single modality, image and LiDAR branch. We apply the pre-trained parameters on the final tracking model using the fused features from different modalities.
    \item The experiments show that our method can perform well on the KITTI benchmark \cite{geiger2012we}. Furthermore, the ablation studies show that multi-scale interactive feature fusion and pre-training on a single modality are beneficial.
\end{itemize}

\section{Related Work}

\subsection{2D and 3D Feature Fusion} Fusing the camera and point cloud features has been proven to improve both accuracy and robustness in 3D object detection and tracking tasks \cite{yoo20203d,mmmot}. 

 Zhang \textit{et al}.\cite{mmmot} designed a multi-modality multi-object tracking framework. Following the paradigm of tracking by detection, the image and point cloud features are extracted separately at first, then through an attention-based fusion module, two single modalities are used to compute a fused modality, which offer a significant performance gain over the single modality.
GNN3DMOT \cite{GNN3DMOT} achieves the fusion of image feature and LiDAR feature through a novel mechanism based on graph neural networks, which benefits the distinguishing feature learning for multi-object learning. EagerMOT \cite{EagerMOT} proposed a novel two-stage data association process, which fuse object detection results from different sensors. Yoo \textit{et al}. \cite{yoo20203d} proposed an approach named auto-calibrated projection method for fusing camera and LiDAR features to achieve better performance in 3D object detection, through which high-precision correspondence  between camera and LiDAR feature map in the BEV domain can be achieved.

\subsection{Multiple Object Tracking}
Many recent works \cite{mmmot,breitenstein2010online} concerning multiple object tracking follow the tracking by detection paradigm, where a detector and a tracker is applied respectively. Firstly, a detector is used to obtain the object detection results in two consecutive frames in a video sequence to be tracked. The detector can be either 2D based on image or 3D based on point cloud. Then the set of detections is fed to the tracker, which will perform the data association process. During the data association process, the detected object will be assigned to an existing trajectory or a newly created empty trajectory through a birth/death check mechanism.
A lot of approaches are used for solving the data association problem, such as Kalman filter \cite{SORT}, Hungarian algorithm\cite{BeyondPixels,SORT}, etc. A linear programming framework can be used to solve these problems \cite{DNF}, which is widely used in the applications of computer vision for solving object matching problems.

Other works \cite{2019MOTS,2017Detect} try to complete the two tasks, detecting and tracking, at the same time. This approach enables the two tasks to benefit each other while learning. CenterTrack\cite{CenterTrack} proposed a novel tracking framework which estimates the object motion of the current frame by performing detection on an image pair and combining the object detection results of the previous frame.

\subsection{Deep Learning on Point Cloud} Point cloud, as an important source of 3D information, is crucial in the field of robust object detection \cite{Multi-Sensor}. Many works have been done to exploit the feature learning of point cloud.

Charles \textit{et al}. PointNet \cite{PointNet} designed a novel framework for processing unstructured point cloud, without requirements for the highly regular input format. Based on PointNet, PointNet++ \cite{PointNet++} is proposed, which is a hierarchical neural network by grouping the local features into larger units to obtain high-level features.
In PointNet++\cite{PointNet++}, two set abstraction layers are proposed, which can combine features from different local point densities. In our work, we applied PointNet++ for the feature extraction of point cloud in each frame, with a different number of points selected due to different density of points in each object detection.

\section{Interactive Feature Fusion for Multi-object Tracking}

\subsection{Problem Formulation}
\begin{figure}[t]
    \centering
    \includegraphics[width=0.50\textwidth]{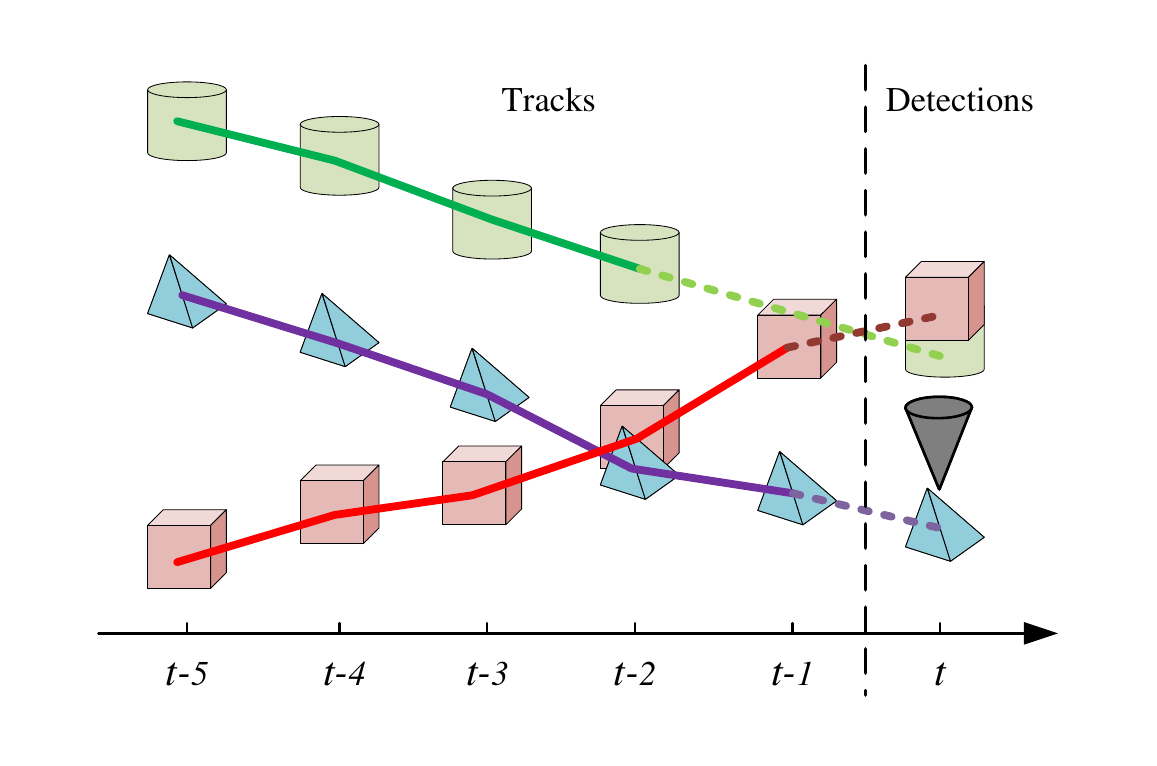}
    \vspace{-6mm}
    \caption{Each shape denotes a detected object. The objective of MOT is to establish the correct connection between tracks and detections, even with false detection.}
    \label{fig:problem_def}
\end{figure}

The objective of Multiple object tracking (MOT) is to assign a consistent and unique ID for the same object in a continuous video sequence. It can be formulated as follows: given a set of object detections $\mathcal{O} = \{o_1, o_2, \dots o_t, \dots, o_T \}$, where $o_t$ denotes the detected objects in frame $t$ and $T$ is length of the video sequence. Each object is represented by a bounding box, $o_t = \{ b_{t1}, b_{t2}, \dots b_{tn}, \dots b_{tN}\}$, where $N$ is the number of objects in frame $t$. A tracklet at time $t-1$ is defined as a set of object detections  in different frames before $t$, $\mathcal{T}_{t-1} = \{\tau_1, \tau_2, \dots, \tau_{t-1} \} $, where $\tau_{t-1}$ is either empty or $\tau_{t-1} \in o_{t-1}$. For multiple object tracking, the goal is to establish the connect between $\mathcal{T}_{t-1}$ and $o_t$. As shown in the Figure \ref{fig:problem_def}, a successful tracking process is to assign an correct ID to each object in frame $t$, based on their similarity to the tracklets before $t$, even with occlusions and false detection.

\begin{figure*}[t]
    \centering
    \includegraphics[width=1.00\textwidth]{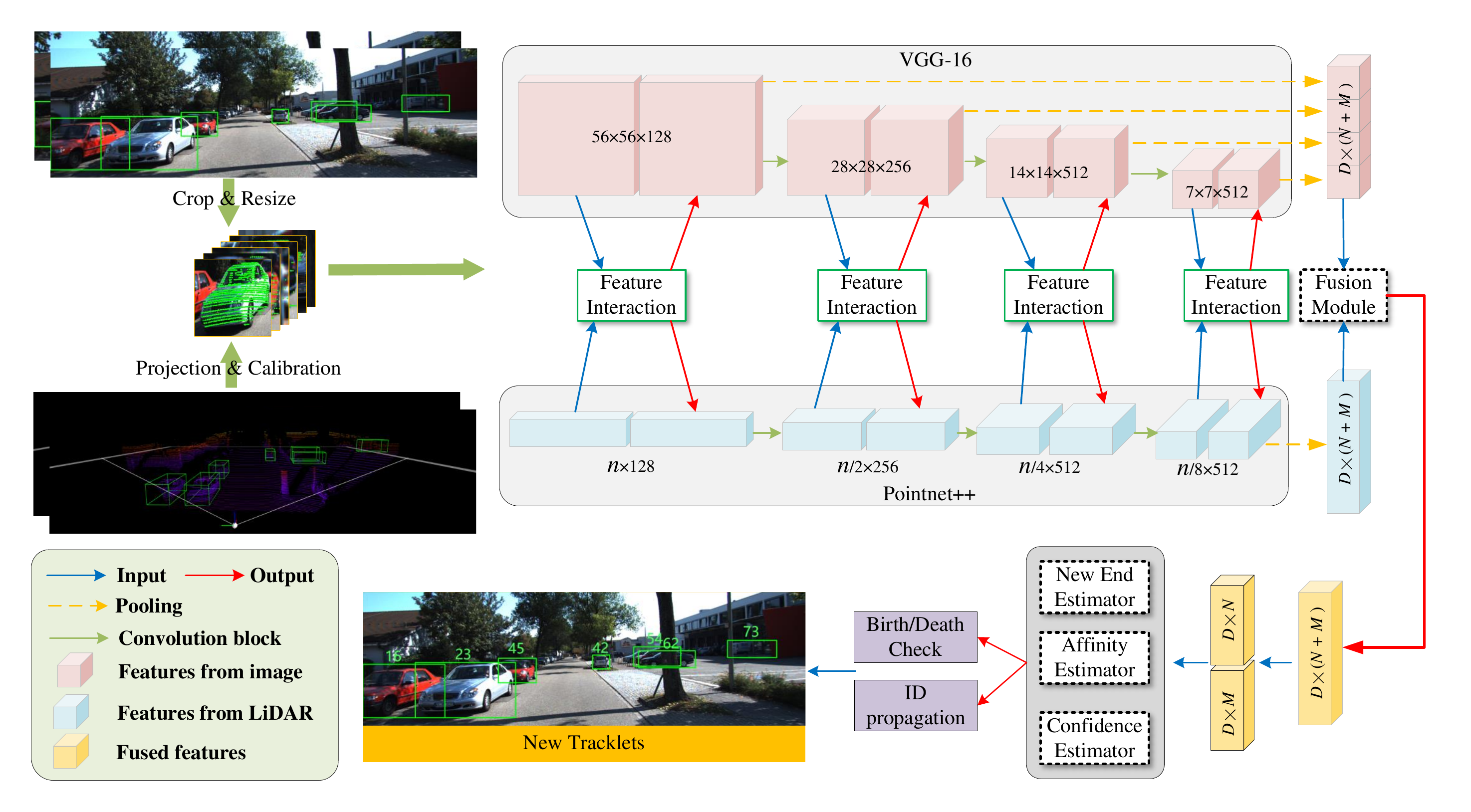}
    \vspace{-8mm}
    \caption{Structure overview of the proposed multi-scale interactive fusion method. The features are extracted by two backbones, VGG-Net and PointNet++, during which a feature interaction module is proposed. The feature interaction between two modalities will be performed for four times, same as the number of layers of the backbone. Then the image and LiDAR features are fed to a fusion module, forming fused features. Next, adjacency estimator will predict the scores based on the fused features. After birth/death check and ID propagation, we can obtain new tracklets.}
    \label{fig:struct}
\end{figure*}

\subsection{Structure Overview} We illustrate the complete structure of our method in Figure \ref{fig:struct}. The whole structure consists of four components. First, we obtain a set of object detections as input. Next, the detection results of 2D image  and 3D point cloud are passed to the second stage to perform feature extraction, respectively. During this stage of feature extraction, the feature interaction module will realize the interactive multi-scale feature fusion between different modalities, image and point cloud, after each layer of the feature extraction network obtaining the deeper features. For the second stage, we use VGG-16\cite{simonyan2014very} and PointNet++ \cite{PointNet++} as the backbones for feature extraction of the image and point cloud, respectively. Both 2D and 3D features will be fused interactively at different levels and updated. The updated features are then passed to the next layer of backbone for further feature extraction and following interactive feature fusion. After obtaining the deep representation of image and point cloud through the two aforementioned backbones, the multi-scale features from two modalities are fed to a final fusion module, which will merge two sets of features into one and achieve final feature fusion. 

In order to train the network in a end-to-end manner, the extracted and fused features will be passed to the adjacency estimation \cite{mmmot}, consisting of three different estimators, start-end estimator, affinity estimator and confidence estimator. The start-end estimator predicts whether an object detection is the beginning  or end of a tracklet, the affinity estimator predicts the linking score between $M$ tracklets in frame $t-1$ and $N$ detections in frame $t$, and the confidence estimator predicts whether an object is false detection to avoid mismatching problems. Following \cite{DNF,frossard2018end}, we map the ID assignment problem to a cost-flow problem and attempt to find an optimal solution through linear programming. After the ID propagation and Birth/Death check procedure, existing tracklets will be updated or deleted and new tracklets will be created according to the predicted scores from the three estimator listed above.

\subsection{Interactive Feature Fusion}
\subsubsection{Projection and Calibration}
\label{subsubsection:pc}
Inspired by the idea of the auto-calibrated projection method in 3D-CVF \cite{yoo20203d} and the set abstraction module from PointNet++ \cite{PointNet++}, we come up with a novel method to better fuse the features from different modalities based on their position relation in physical spaces.

The LiDAR points are first projected on the full image plane through calibration matrix. The 2D detection results are represented by a 2D bounding box. In our method, each image patch within a bounding box is cropped from the full image, then resized to the shape of $224 \times 224$, taken as input to the VGG-Net backbone. As a result of the resizing process, the projected LiDAR points need to be calibrated according to the size of the resized picture. The project matrix has been changed, which means the point clouds need to be re-projected if the projection plane changed.

The coordinates of the projected points on full image is denoted as $(u,v)$, and the position of the bounding box is $x_{1}, y_{1}, x_{2}, y_{2}$, where $x_{1} < x_{2}, y_{1} < y_{2}$. Therefore, the coordinates of projected points on the cropped picture (not resized yet) can be expressed as $(u^{'},v^{'})=(u-x_{1},v-y_{1})$, then the resized position $(u^{''},v^{''})$ can be calculated using the zooming principle in the $x$ and $y$ axes.
\begin{equation}
\label{eq:uv}
(u^{''},v^{''})=(\frac{W}{\Delta x} u^{'} ,\frac{H}{\Delta y} v^{'}),    
\end{equation}
where $\Delta x = x_{2} - x_{1},\Delta y = y_{2} - y_{1} $, and $W, H$ is the width and height of the resized image, respectively. If equation (\ref{eq:uv}) is written as a matrix, \textit{M-matrix}, it is shown in equation (\ref{eq:mmat}). It's noteworthy that the \textit{M-matrix} for each image patch in a bounding box is usually different, because the image patches can vary in size, both width and height. On the other hand, the projected points within the same bounding box can be calibrated via multiplying the coordinates by the same \textit{M-matrix}. Through this re-scaled projection method, the point clouds can be projected to the right place no matter how the size of image patch varies and satisfactory correspondence between the position of point clouds and the scaled image patch can be obtained.
\begin{equation}
\label{eq:mmat}
\begin{pmatrix}
u^{''} \\[10pt] v^{''} \\[10pt] 1
\end{pmatrix}
=
\begin{pmatrix}
\dfrac {W} {\Delta x} & 0 & -\dfrac {Hx_{1}} {\Delta x} \\[10pt]
0  & \dfrac {W} {\Delta y}  & -\dfrac {Hy_{1}} {\Delta y}  \\[10pt]
0 & 0 & 1
\end{pmatrix}
\begin{pmatrix}
u \\[10pt] v \\[10pt] 1
\end{pmatrix}
\triangleq{\textbf{$M \cdot P_{uv}$}}.
\end{equation}

\begin{figure*}[t]
    \centering
    \includegraphics[width=1.00\textwidth]{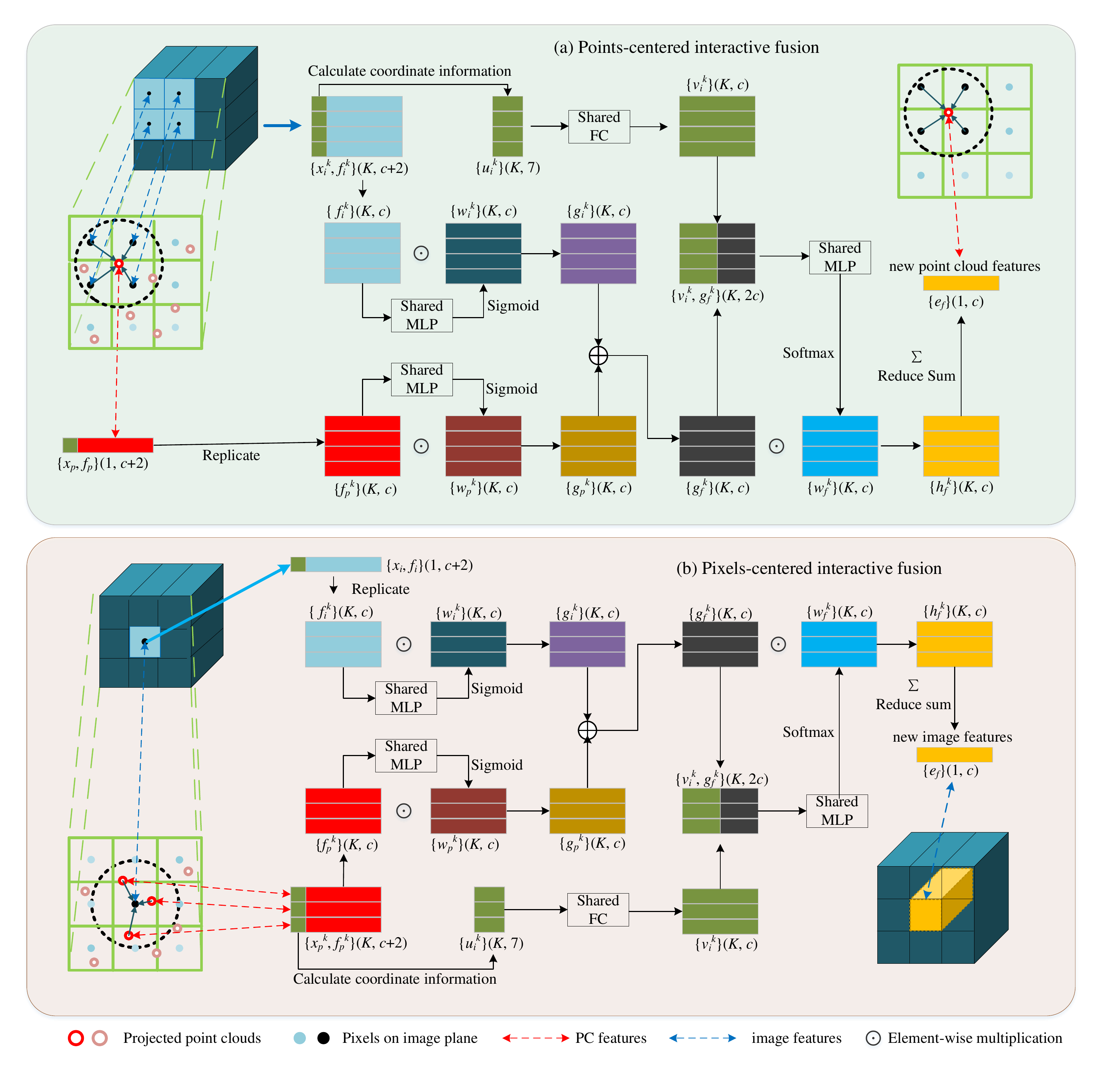}
    \vspace{-12mm}
    \caption{Details of the proposed interactive feature fusion. (a) The feature of a point is updated through the fusion with the features of $K$ neighborhood pixels. (b) The feature of a pixel is updated through the fusion with the features of $K$ neighborhood points.}
    \label{fig:fuse}
\end{figure*}

\subsubsection{Points-centered Interactive Fusion}
\label{subsubsection:points-centered}
In this module, we set the projected LiDAR points as the center whose features are updated by grouping the features of the neighborhood image pixels. As shown in Figure \ref{fig:fuse} (a), first, the LiDAR points are projected to image plane, using the projection technique described in Section \ref{subsubsection:pc}. Take one center point as an example, the multi-scale interactive fusion procedure can formulated as follows.  

Let $\{x_p, f_p\}$ denote the position and features of the center point and $\{x_i,f_i\}$ denote that of pixels. $K$ nearest neighboring pixels, whose features are represented by $f_i^k$, are grouped for a center point based on their distances on the 2D image plane. After multiple times of feature extraction, the number of points decreases. Without a limited radius, far-away pixels tend to be grouped, which harms the extraction of distinguishing features. Therefore, instead of purely select $K$ nearest pixels regardless of the distance, we additionally set a searching radius $r$ for the KNN algorithm to filter out the neighbors that are too far away from the center point. 
The shape of $f_i^k$ is $K \times D$, where $D$ is number of feature channel. Next, the features $f_p$ of the center point are replicated for $K$ times, generating $f_p^k$, whose shape is also $K \times D$. 

Given the grouped pixels and center LiDAR point, an attention mechanism is proposed to effectively fuse their features to obtain better representations of the local structures. First, two separate Multi-Layer Perceptrons (MLP) are used to obtain a self-attention weights from the original features $f_p^k$ and $f_i^k$. The attentive weights are: 
\begin{equation}
w_i^k = \text{Sigmoid}(\text{MLP}(f_i^k)),
\end{equation}
\begin{equation}
w_p^k =\text{Sigmoid}(\text{MLP}(f_p^k)).
\end{equation}

By multiplying the features with their corresponding attentive weights, we can get the weighted features from two modalities, $g_i^k$ and $g_p^k$. The fused feature $g_f^k$ is the sum of the element-wise products of $f_i^k, w_i^k$ and $f_p^k, w_p^k$ calculated as Equation (\ref{eq:sum}), where $\odot$ indicates element-wise multiplication. Through the attention weights,  different degrees of importance can be assigned to different pixels. 
\begin{equation}
\label{eq:sum}
g_f^k = \frac{g_i^k + g_p^k}{w_i^k + w_p^k} = \frac{f_i^k \odot w_i^k + f_p^k \odot w_p^k}{w_i^k + w_p^k}.
\end{equation}

 In addition to feature correlation, the positional relationship between different modality is also considered in feature fusion. The coordinate information $u_i^k \in R^{k \times 7}$ containing position relation between points and pixels is calculated as Equation (\ref{eq:pos}), including the original coordinates, the differences of coordinates, and the euclidean distances. We believe that these three components contain more abundant and comprehensive information than the original coordinates.
\begin{equation}\label{eq:pos}
u_i^k = x_i^k \oplus y_i^k \oplus (x_i^k - y_i^k) \oplus \lVert x_i^k - y_i^k \rVert,
\end{equation}
where $\oplus$ denotes concatenation operation between tensors, and $\lVert \cdot \rVert$ denotes euclidean distance between two points.

For the second stage, our method attempts to combine the position information $u_i^k$  with the fused feature $g_f^k$ of the first stage to generate an attention map. A fully connected layer (FC) is used as an encoder for position information, making the shape of encoded position information exactly same as the fused feature. Then, a MLP is used to process the concatenated information, generating the attentive weights of the second stage $w_f^k \in R^{K \times D}$.
\begin{equation}
w_f^k = \text{Softmax}(\text{MLP}(g_f^k \oplus \text{FC}(u_i^k))).
\end{equation}
Lastly, by summing up the weighted fused features in the $K$ dimension, we get the final fused features $e_f$ of the point cloud located at $x_p$.
\begin{equation}
 e_f = \sum_{k=1}^K g_f^k \odot w_f^k.   
\end{equation}

The original point cloud feature is then replaced by the fused feature and fed to the next layer of backbone for further feature extraction.

\subsubsection{Pixels-centered Interactive Fusion}
In a manner similar to the points-centered interactive fusion process described in Section \ref{subsubsection:points-centered}, the position of pixels $x_i$ is used to search $K$ nearest neighbors $x_p^k$ among points projected on the image plane, whose features are denoted as $f_p^k$. 

After multiplying $f_p^k$ by the weights $w_p^k \in R^{K \times D}$, which is obtained from a shared MLP, we can obtain the grouped point cloud features $g_p^k \in R^{K \times D}$. In the following equations,  $\odot$ indicates element-wise multiplication.
\begin{equation}
g_p^k = f_p^k \odot \text{Sigmoid}(\text{MLP}(f_p^k)).  
\end{equation}
Likewise, the image features of center pixel are extracted as follows:
\begin{equation}
g_i^k = f_i^k \odot \text{Sigmoid}(\text{MLP}(f_i^k)).    
\end{equation}

Next, the features from different modalities are added up, generating $g_f^k \in R^{K \times D}$, and then concatenated with position information $v_i^k$ encoded by a fully connected layer (FC) from the calculated coordinate information $u_i^k$ same as Equation (\ref{eq:pos}). After feeding the concatenated information to a shared MLP for processing, the attentive weights $w_f^k$ of the second stage can be obtained as follows:
\begin{equation}
w_f^k = \text{Softmax}(\text{MLP}(g_f^k \oplus v_i^k)),
\end{equation}
where $\oplus$ denotes concatenation between tensors, and $\odot$ denotes element-wise multiplication. At last, the new image feature $e_f \in R^{D}$  is obtained from the weighted sum of the fused features $g_f^k$ in the $K$ dimension, as follows:
\begin{equation}
    e_f = \sum_{k=1}^K g_f^k \odot w_f^k.
\end{equation}

To summarize, the interactive fusion for pixels is quite similar to the points-centered interactive fusion, with the exception of the specified $K$ and the radius for KNN, which are different because the density of pixels is significantly larger than the density of points.

\subsubsection{Further Feature Fusion}
During the feature extraction process, we achieve feature interaction between image domain and point cloud domain. 
Given the image and point cloud features obtained from the backbone, we merge features from two different modalities into one for the following predictions.  The camera and LiDAR features are fused with a attention mechanism \cite{mmmot} shown as follows:
\begin{align}
   G_{c} & = \text{Sigmoid}(\text{MLP}(F_{c})), \\
   G_{l} &= \text{Sigmoid}(\text{MLP}(F_{l})),\\
   F_{f} & = \frac{G_c \odot F_c + G_l \odot F_l}{G_c + G_l}.
\end{align}

where $\odot$ denotes element-wise multiplication, and $F_c, F_l, F_f \in R ^ {D}$ denotes the image, LiDAR and fused features respectively. Through this fusion module, the features from different modalities are assigned different degrees of significance using the learned weights $G_c$ and $G_l$, which contributes to  deal with the circumstances where one sensor fails or not reliable enough.

\subsubsection{Adjacency Estimation}
In order to train the network in an end-to-end manner, we adopt the adjacency matrix learning from \cite{mmmot}, consisting of three estimators. 

The new/end estimator predicts whether an object detection in frame $t$ is the beginning or end of a tracklet. If the estimator determines that a detection is a start point, a new tracklet will be created and a new ID is assigned to the detection. If one detection is determined as an end, it will be appended to an existing tracklet, and that tracklet will not be participate in the next process of data association between frame $t$ and $t+1$.  The confidence estimator predicts whether an object detection in frame $t$ is a false detection, which helps avoid mis-matching problems. For the adjacency estimator, it predicts the linking score $s_{ij}$, determining whether there is connection between the $i$-th tracklet of the $M$ tracklets in frame $t-1$ and the $j$-th detection of the $N$ object detections in frame $t$.

\subsection{Pre-training on Different Modalities}
Pre-training has been proven to be quite effective in some previous work \cite{hendrycks2019using}. It usually consists of two steps, 1) pre-training the model on abundant data concerning upstream task, 2) fine-tuning on downstream specific tasks. However, \cite{Pre-train} points out that there exists a gap between the pre-training and fine-tuning if the task is different due to the slightly different optimization objectives, but all of these above problems will not influence our work in this paper.  Upstream and downstream tasks are not included in our pre-training, instead, only multi-object tracking task is performed, which ensures that there are no gaps in optimization objectives between the upstream task and downstream task during the training process. 

First, we use only image feature for the aforementioned adjacency matrix learning and the subsequent tracking task. Secondly, we do the similar process based on the feature extracted from point clouds. From the two steps, we can obtain two pre-trained feature extractors for different modalities. Finally, we put the two pre-trained backbones into the complete tracking framework, which utilizes both the image and the point cloud features to generate fused features to further aid the multi-modality multi-object tracking task. 

\begin{table*}[htpb]
  \caption{Comparison on the testing datasets of KITTI}
  \label{table:res}
  \centering
  \begin{tabular}{l|ccccccccccc}
    \toprule
    Method & MOTA$\uparrow$ & MOTP$\uparrow$  & ID-s$\downarrow$ & HOTA$\uparrow$ & DetA $\uparrow$ & AssA $\uparrow$ &  FN$\downarrow$ & FP$\downarrow$  & Frag$\downarrow$ & MT$\uparrow$ & ML$\downarrow$\\
    \midrule
    DSM \cite{frossard2018end} &  73.94 & 83.5  & 939  & 60.05 & 64.09 & 57.18 & 637  & 7388 & 737      & 59.38 & 8.46  \\
    extraCK \cite{gunduz2018lightweight} & 79.29 & 82.06 & 520  & 59.76 & 65.18 & 55.47 & 675  & 5929 & 750 & 62.31 & 5.85  \\
    MOTBeyondPixels \cite{BeyondPixels} &  82.68 & \textbf{85.50}  & 934  & 63.75 & 72.87 & 56.4  & 741  & \textbf{4283} & 581 & 72.61 & 2.92  \\
    3D-CNN/PMBM \cite{PMBM} & 79.23 & 81.58 & 485  & 59.12 & 65.43 & 54.28 & 1024 & 5634 & 554      & 62.77 & 6.46  \\
    JCSTD \cite{Tian2019MOT} & 80.24 & 81.85 & \textbf{173}  & 65.94 & 65.37 & 67.03 & \textbf{405}  & 6217 & 700 & 57.08 & 7.85  \\
    IMMDP \cite{Ren2015NIPS} & 82.75 & 82.78 & 211  & \textbf{68.66} & 68.02 & \textbf{69.76} & 422  & 5300 & \textbf{201} & 60.31 & 12.15 \\
    mmMOT \cite{mmmot} & 83.23 & 85.03 & 733  & 62.05 & \textbf{72.29} & 54.02 & 752  & 4284 & 570  & 72.92 & 2.92  \\

    \midrule
    Ours & \textbf{84.24} & 85.03 & 415 & 62.69 &	72.22 &	55.20 &  713 & 4292 &  568 & \textbf{72.97} & \textbf{2.77}\\
    \bottomrule
  \end{tabular}
\end{table*}

\begin{figure*}[htpb]
    \centering
    \includegraphics[width=1\textwidth]{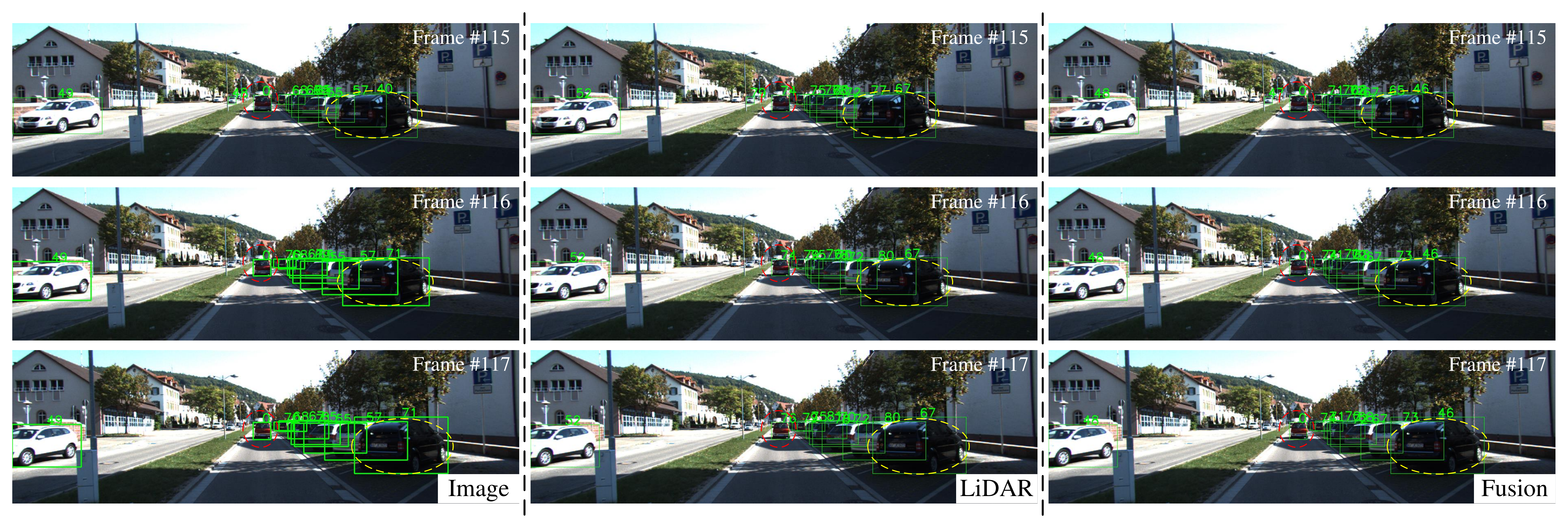}
    \vspace{-8mm}
    \caption{Qualitative results of our multi-object tracking method. We illustrate a comparison between tracking results using different modalities. The first column is the tracking results with only images as inputs, the second column is that with only LiDAR point clouds as inputs, and the third column is that with both as inputs.} 
    \label{fig:qual}
\end{figure*}

\section{Experiments}
\subsection{Settings}
\subsubsection{Dataset}
The KITTI benchmark \cite{geiger2012we} is a well-known benchmark for applications on autonomous driving and related technologies.
All of our experiments is conducted on the KITTI MOT dataset, which consists of 21 training sequences and 29 testing sequences. Since the official dataset does not provide a training/validation split, following \cite{mmmot}, we choose sequences 00, 02, 05, 07, 10, 11, 14, 16, 17, 18, 19 as the validation dataset and take the rest as training dataset.
\subsubsection{Evaluation Metrics} 
\label{section:eval}We use standard MOT metrics, CLEAR MOT \cite{bernardin2008evaluating} (including Multiple Object Tracking Accuracy (MOTA), Multiple Object Tracking Precision (MOTP),  ID
Switches (IDS), Fragmentation (FRAG), False Negatives (FN) and False Positives (FP))  and Mostly Tracked
targets (MT) / Mostly Lost targets (ML)\cite{li2009learning}, to evaluate the performance of our method on KITTI Benchmark.  Apart from the metrics mentioned above, we also adopt Higher Order Tracking Accuracy (HOTA), Detection Accuracy
(DetA) and Association
Accuracy (AssA) \cite{luiten2021hota}, which enables better understanding of tracking behavior than previous metrics.
\subsection{Implementation Details}
Our models are trained on a TITAN RTX, and the batch size we use is 1. We use ADAM as our optimizer with a learning rate of $3\times 10^{-6}$. 

\subsubsection{Object Detection}
For evaluation on the KITTI testing set,  we use the 3D detection results produced by PointPillars \cite{lang2019pointpillars} and 2D detection results obtained from RRC-Net \cite{ren2017accurate}. 

\subsubsection{Feature Extraction}
We use VGG-Net\cite{simonyan2014very} and PointNet++\cite{PointNet++} as the backbone for feature extraction of image and point cloud, respectively.

For image domain, skip pooling layers\cite{simonyan2014very} are applied to the output feature of each layer of the VGG-Net backbone, whose number of channels is 128, 256, 512, 512 in turn. After the process of skip pooling layers, we can obtain four vectors with an identical length of 128. Next, the four vectors are concatenated together, forming a vector whose length is 512, then passing to the fusion module.

For LiDAR branch, we use PointNet++\cite{PointNet++} as our backbone. We take the point clouds filter by the frustum projected by the 2D bounding box as input. Point clouds from different object detections are combined together, forming a $n \times 3$ vector, where $n = n_1 + n_2 + n_3 + \cdots + n_l $. $n_1, n_2, \cdots n_l$ denote the number of points in each frustum and $l$ denotes the number of frustums. The point clouds are then passed to the PointNet++\cite{PointNet++} for feature extraction. As the number of points is decreasing from $n$ to $n/2, n/4, n/8$, the number of feature channels is increasing from 128 to 256, 512, 512. Finally, we obtain a deeper representation of point cloud and feed it to the fusion module.
\subsection{Results}

We compare our method with other published state-of-art methods \cite{frossard2018end,gunduz2018lightweight,BeyondPixels,PMBM,Tian2019MOT,Ren2015NIPS,mmmot} on the KITTI testing dataset with the evaluation metrics introduced in Section \ref{section:eval}.

As Table \ref{table:res} shows, our method obtains the best performance in terms of MOTA, which is a comprehensive metric in evaluating the performance in multi-object tracking. We believe that the proposed multi-scale interactive feature fusion module in this paper contributes greatly to this improvement. Our method also outperforms the baseline, mmMOT\cite{mmmot}, with fewer ID-s even using the same detection results. Because of the interaction between points and pixels during feature extraction process, the extracted features are more distinguishing, alleviating the mismatching problems.

Figure \ref{fig:qual} shows the visualization results of our method under three different modality settings. It can be observed in the first column that the image-based tracking failed because of large displacement of objects in adjacent frames and occlusion between crowded objects, which causes the mismatching problems. For the LiDAR branch in the second column, the sparsity of point cloud of objects in the long distance leads to the failure of ID propagation while tracking. Focusing on the car annotated by the red circle and the yellow circle, in the last column, it can be observed that the fusion-based tracking outperforms the other two single-modality methods.

\begin{table}[t]
  \caption{Ablation study of interactive feature fusion layers}
  \label{table:ablation1}
  \centering
  \begin{tabular}{cccccc}
  \toprule
    Layer 1 & Layer 2 & Layer 3 & Layer 4 & MOTA$\uparrow$ & ID-s$\downarrow$  \\
    \midrule
    $\quad$ & $\quad$ & $\quad$ & $\quad$ & 90.92 & 194 \\
    $\quad $ & $\quad$ & $\quad$ & $\checkmark$ & 91.19 & 165 \\
    $\quad$ & $\quad $ & $\checkmark$ & $\checkmark$ & 91.11 & 173\\
    $\quad$ & $\checkmark$ & $\checkmark$ & $\checkmark$ & 91.12 & 173 \\
    $\checkmark$ &  $\checkmark$ & $\checkmark$ & $\checkmark$ &91.56 & 124 \\
    \bottomrule
  \end{tabular}
\end{table}

\begin{table}[t]
  \caption{Ablation study of Pre-training on Sinle and Multiple Modalities}
  \label{table:ablation2}
  \centering
    \begin{tabular}{c|c|ccc}
      \toprule
      Pre-trained model & Modality & MOTA$\uparrow$  & FP $\downarrow$ & FN$\downarrow$ \\
    \midrule
      \multirow{3}{*}{No pre-training} & Image & 78.40 &  951 & 1387\\
       & Point Cloud & 77.15  & 942 & 1388 \\
       & Fusion & 78.16  & 941 & 1402 \\
      \midrule
      
      \multirow{3}{*}{VGG-Net} & Image & 78.52 &  782 & 1352\\
       & Point Cloud & 76.37  & 942 & 1393 \\
       & Fusion & 78.60  & 951 & 1387 \\
      \midrule
      \multirow{3}{*}{PointNet++} & Image & 78.27  & 945 & 1411\\
       & Point Cloud & 77.26  & 906 & 1526 \\
       & Fusion & 78.77  & 897 & 1431 \\
      \midrule
      VGG-Net & Image & 78.60  & 951 & 1387\\
      $\&$ & Point Cloud & 77.23  & 929 & 1393 \\
      PointNet++  & Fusion & 79.22  & 935 & 1389\\

       \bottomrule
    \end{tabular}
\end{table}

\subsection{Ablation Study}
To evaluate the effectiveness of our method, we conduct an ablation study on the validation dataset of KITTI\cite{geiger2012we}.
\subsubsection{Interactive Feature Fusion Module}
To evaluate the effectiveness of our proposed multi-scale interactive feature fusion module, we compare the tracking performances with or without the four interactive feature fusion layers. From the top and bottom row in Table \ref{table:ablation1}, it can be observed that using the interactive fusion module can improve the MOTA greatly over the baseline. Then we attempt to remove a few layers to see how much one single layer can affect the final results. From the second row to the fifth row of Table \ref{table:ablation1}, it can be observed that each layer contributes to the improvement on MOTA to some different degree.

\subsubsection{Pre-training on Single and Multiple Modalities}
We compare the effectiveness of pre-training on VGG-Net\cite{simonyan2014very}, PointNet++\cite{PointNet++}. We present the experimental results with different modalities, image, LiDAR point cloud and both of them, under the different pre-training settings. 
In Table \ref{table:ablation2}, it can be observed that pre-training on a single modality can contribute to the improvements of MOTA. Specifically, pre-training on VGG-Net can benefit the results with the inputs of images, and  pre-training on PointNet++ can benefit the results with the inputs of point clouds. When pre-training on both VGG-net and PointNet++ is conducted, the best performance of tracking based on fusion modality is achieved. 
Specifically, these two backbones can benefit each other to report a MOTA score of 79.22 using fused features, outperforming over other pre-training settings. 

\begin{figure}[t]
    \centering
    \includegraphics[width=0.5\textwidth]{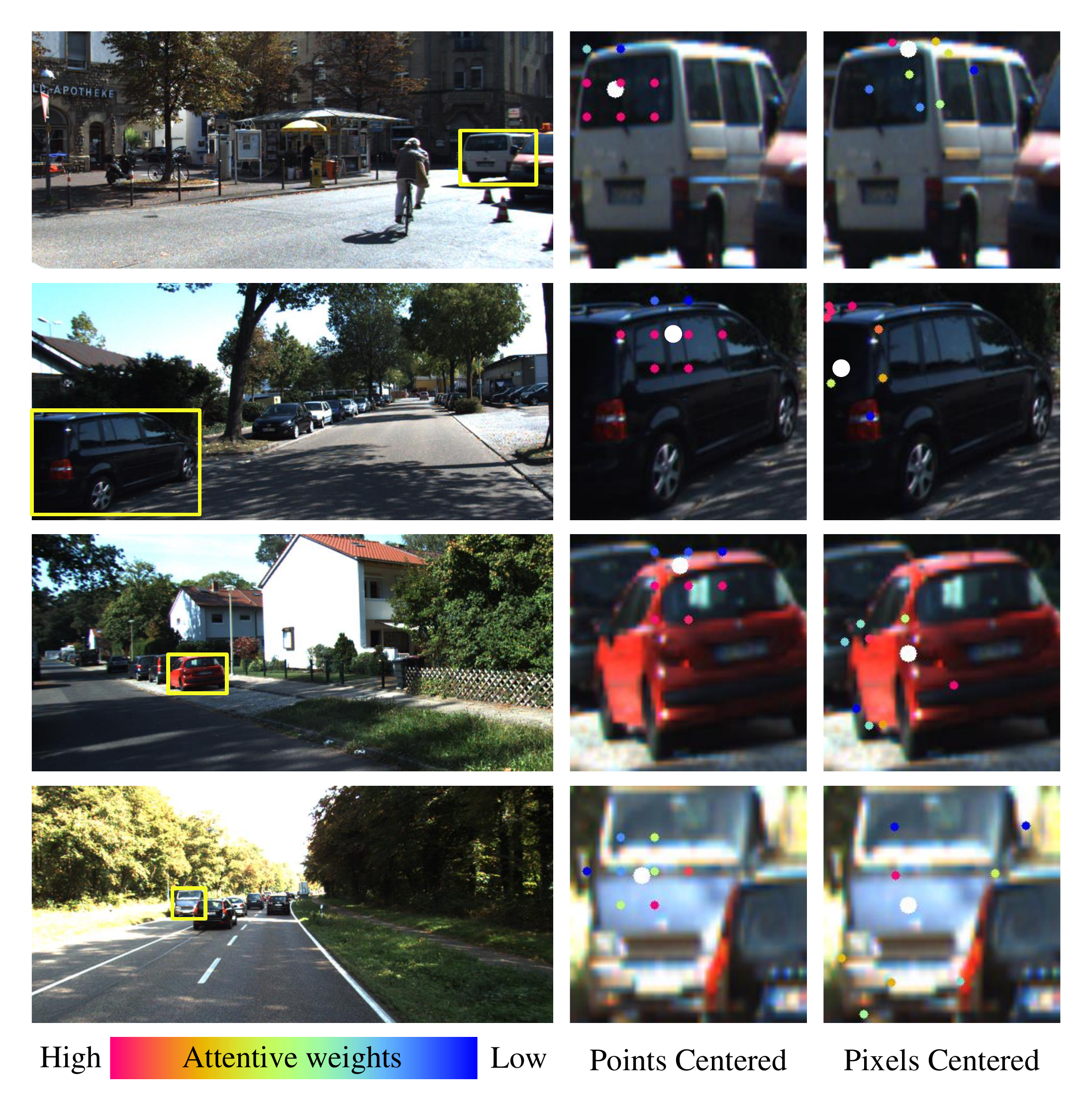}
    \vspace{-5mm}
    \caption{The visualization of attentive weights during interactive fusion. The first column is the bounding box, the second column shows the weights of points-centered fusion method, and the third column shows the weights of pixels-centered fusion method.}
    \label{fig:visualize}
\end{figure}

\subsection{Visualization of the Attentive Fusion}
In Figure \ref{fig:visualize}, we visualize the attentive weights of the two different interactive fusion methods, where the white dot denotes the center point. In the second column, the grouped pixels outside the vehicle (blue) possesses a lower weight value than pixels inside the vehicle (red). In the third column, our method tends to attach higher weight value to the points, which possess similar textures to the center pixel.

\section{Conclusion}
\label{sec:conclusion}
In this paper, we proposed a novel approach named Multi-Scale Interactive Feature Fusion for the fusion process of features from two different modalities, RGB images and LiDAR point cloud. The proposed method can not only achieve better performance in fusing the features from multiple sensors, but also fully exploit the positional relationship between points of point cloud and pixels of images by introducing PointNet++ as the backbone for feature extraction of LiDAR point cloud. We also propose the pre-training method on each single modality for the fusion-based multi-object tracking. Extensive ablation studies  demonstrate the effectiveness of our proposed methods.  We believe that the proposed fusion method will enlighten other researchers on exploring the spatial relation between points and pixels to achieve better fusion in the field of multi-modality perception.

\bibliographystyle{IEEEtran}  
\bibliography{IEEEabrv,ref} 

\begin{thebibliography}{10}
\providecommand{\url}[1]{#1}
\csname url@samestyle\endcsname
\providecommand{\newblock}{\relax}
\providecommand{\bibinfo}[2]{#2}
\providecommand{\BIBentrySTDinterwordspacing}{\spaceskip=0pt\relax}
\providecommand{\BIBentryALTinterwordstretchfactor}{4}
\providecommand{\BIBentryALTinterwordspacing}{\spaceskip=\fontdimen2\font plus
\BIBentryALTinterwordstretchfactor\fontdimen3\font minus
  \fontdimen4\font\relax}
\providecommand{\BIBforeignlanguage}[2]{{%
\expandafter\ifx\csname l@#1\endcsname\relax
\typeout{** WARNING: IEEEtran.bst: No hyphenation pattern has been}%
\typeout{** loaded for the language `#1'. Using the pattern for}%
\typeout{** the default language instead.}%
\else
\language=\csname l@#1\endcsname
\fi
#2}}
\providecommand{\BIBdecl}{\relax}
\BIBdecl

\bibitem{xiang2020end}
J.~Xiang, G.~Xu, C.~Ma, and J.~Hou, ``End-to-end learning deep crf models for
  multi-object tracking deep crf models,'' \emph{IEEE Trans. Circuits Syst.
  Video Technol.}, vol.~31, no.~1, pp. 275--288, 2020.

\bibitem{zhou2018deep}
H.~Zhou, W.~Ouyang, J.~Cheng, X.~Wang, and H.~Li, ``Deep continuous conditional
  random fields with asymmetric inter-object constraints for online
  multi-object tracking,'' \emph{IEEE Trans. Circuits Syst. Video Technol.},
  vol.~29, no.~4, pp. 1011--1022, 2018.

\bibitem{BeyondPixels}
S.~Sharma, J.~A. Ansari, J.~K. Murthy, and K.~M. Krishna, ``Beyond pixels:
  Leveraging geometry and shape cues for online multi-object tracking,'' in
  \emph{Proc. IEEE Int. Conf. Robot. Autom. (ICRA)}, 2018, pp. 3508--3515.

\bibitem{bergmann2019tracking}
P.~Bergmann, T.~Meinhardt, and L.~Leal-Taixe, ``Tracking without bells and
  whistles,'' in \emph{Proc. IEEE Int. Conf. Comput. Vis. (ICCV)}, 2019, pp.
  941--951.

\bibitem{2019MOTS}
P.~Voigtlaender, M.~Krause, A.~Osep, J.~Luiten, and B.~Leibe, ``Mots:
  Multi-object tracking and segmentation,'' in \emph{Proc. IEEE Conf. Comput.
  Vis. Pattern Recognit. (CVPR)}, 2019.

\bibitem{weng20203d}
X.~Weng, J.~Wang, D.~Held, and K.~Kitani, ``3d multi-object tracking: A
  baseline and new evaluation metrics,'' in \emph{IEEE Int. Conf. Intell.
  Robots Syst. (IROS)}, 2020, pp. 10\,359--10\,366.

\bibitem{simonyan2014very}
K.~Simonyan and A.~Zisserman, ``Very deep convolutional networks for
  large-scale image recognition,'' \emph{arXiv preprint arXiv:1409.1556}, 2014.

\bibitem{ResNet}
K.~He, X.~Zhang, S.~Ren, and J.~Sun, ``Deep residual learning for image
  recognition,'' in \emph{Proc. IEEE Conf. Comput. Vis. Pattern Recognit.
  (CVPR)}, 2016, pp. 770--778.

\bibitem{PointNet}
C.~R. Qi, H.~Su, K.~Mo, and L.~J. Guibas, ``Pointnet: Deep learning on point
  sets for 3d classification and segmentation,'' in \emph{Proc. IEEE Conf.
  Comput. Vis. Pattern Recognit. (CVPR)}, 2017, pp. 652--660.

\bibitem{PointNet++}
C.~R. Qi, L.~Yi, H.~Su, and L.~J. Guibas, ``Pointnet++: Deep hierarchical
  feature learning on point sets in a metric space,'' in \emph{Proc. Int. Conf.
  Neural Inform. Process. Syst. (NIPS)}, vol.~30, 2017.

\bibitem{shi2019pointrcnn}
S.~Shi, X.~Wang, and H.~Li, ``Pointrcnn: 3d object proposal generation and
  detection from point cloud,'' in \emph{Proc. IEEE Conf. Comput. Vis. Pattern
  Recognit. (CVPR)}, 2019, pp. 770--779.

\bibitem{CenterTrack}
X.~Zhou, V.~Koltun, and P.~Kr{\"a}henb{\"u}hl, ``Tracking objects as points,''
  in \emph{Proc. Eur. Conf. Comput. Vis. (ECCV)}, 2020, pp. 474--490.

\bibitem{mmmot}
W.~Zhang, H.~Zhou, S.~Sun, Z.~Wang, J.~Shi, and C.~C. Loy, ``Robust
  multi-modality multi-object tracking,'' in \emph{Proc. IEEE Int. Conf.
  Comput. Vis. (ICCV)}, 2019.

\bibitem{GNN3DMOT}
X.~Weng, Y.~Wang, Y.~Man, and K.~M. Kitani, ``Gnn3dmot: Graph neural network
  for 3d multi-object tracking with 2d-3d multi-feature learning,'' in
  \emph{Proc. IEEE Conf. Comput. Vis. Pattern Recognit. (CVPR)}, 2020, pp.
  6499--6508.

\bibitem{EagerMOT}
A.~Kim, A.~O{\v{s}}ep, and L.~Leal-Taix{\'e}, ``Eagermot: 3d multi-object
  tracking via sensor fusion,'' in \emph{Proc. IEEE Int. Conf. Robot. Autom.
  (ICRA)}, 2021, pp. 11\,315--11\,321.

\bibitem{deng2009imagenet}
J.~Deng, W.~Dong, R.~Socher, L.-J. Li, K.~Li, and L.~Fei-Fei, ``Imagenet: A
  large-scale hierarchical image database,'' in \emph{Proc. IEEE Conf. Comput.
  Vis. Pattern Recognit. (CVPR)}, 2009, pp. 248--255.

\bibitem{geiger2012we}
A.~Geiger, P.~Lenz, and R.~Urtasun, ``Are we ready for autonomous driving? the
  kitti vision benchmark suite,'' in \emph{Proc. IEEE Conf. Comput. Vis.
  Pattern Recognit. (CVPR)}, 2012, pp. 3354--3361.

\bibitem{yoo20203d}
J.~H. Yoo, Y.~Kim, J.~Kim, and J.~W. Choi, ``3d-cvf: Generating joint camera
  and lidar features using cross-view spatial feature fusion for 3d object
  detection,'' in \emph{Proc. Eur. Conf. Comput. Vis. (ECCV)}, 2020, pp.
  720--736.

\bibitem{breitenstein2010online}
M.~D. Breitenstein, F.~Reichlin, B.~Leibe, E.~Koller-Meier, and L.~Van~Gool,
  ``Online multiperson tracking-by-detection from a single, uncalibrated
  camera,'' \emph{IEEE Trans. Pattern Anal. Mach. Intell.}, vol.~33, no.~9, pp.
  1820--1833, 2010.

\bibitem{SORT}
A.~Bewley, Z.~Ge, L.~Ott, F.~Ramos, and B.~Upcroft, ``Simple online and
  realtime tracking,'' in \emph{Proc. IEEE Int. Conf. Image Process. (ICIP)},
  2016, pp. 3464--3468.

\bibitem{DNF}
S.~Schulter, P.~Vernaza, W.~Choi, and M.~Chandraker, ``Deep network flow for
  multi-object tracking,'' in \emph{Proc. IEEE Conf. Comput. Vis. Pattern
  Recognit. (CVPR)}, 2017, pp. 6951--6960.

\bibitem{2017Detect}
C.~Feichtenhofer, A.~Pinz, and A.~Zisserman, ``Detect to track and track to
  detect,'' in \emph{Proc. IEEE Int. Conf. Comput. Vis. (ICCV)}, 2017.

\bibitem{Multi-Sensor}
M.~Bai, G.~Mattyus, N.~Homayounfar, S.~Wang, S.~K. Lakshmikanth, and
  R.~Urtasun, ``Deep multi-sensor lane detection,'' in \emph{IEEE Int. Conf.
  Intell. Robots Syst. (IROS)}, 2018, pp. 3102--3109.

\bibitem{frossard2018end}
D.~Frossard and R.~Urtasun, ``End-to-end learning of multi-sensor 3d tracking
  by detection,'' in \emph{Proc. IEEE Int. Conf. Robot. Autom. (ICRA)}, 2018,
  pp. 635--642.

\bibitem{hendrycks2019using}
D.~Hendrycks, K.~Lee, and M.~Mazeika, ``Using pre-training can improve model
  robustness and uncertainty,'' in \emph{Proc. Int. Conf. Mach. Learn. (ICML)},
  2019, pp. 2712--2721.

\bibitem{Pre-train}
Y.~Lu, X.~Jiang, Y.~Fang, and C.~Shi, ``Learning to pre-train graph neural
  networks,'' in \emph{Proc. AAAI Conf. Artif. Intell.}, 2021.

\bibitem{gunduz2018lightweight}
G.~Gunduz and T.~Acarman, ``A lightweight online multiple object vehicle
  tracking method,'' in \emph{Proc. IEEE Intell. Vehicle Symp. (IV)}, 2018, pp.
  427--432.

\bibitem{PMBM}
S.~Scheidegger, J.~Benjaminsson, E.~Rosenberg, A.~Krishnan, and
  K.~Granstr{\"o}m, ``Mono-camera 3d multi-object tracking using deep learning
  detections and pmbm filtering,'' in \emph{Proc. IEEE Intell. Vehicle Symp.
  (IV)}, 2018, pp. 433--440.

\bibitem{Tian2019MOT}
W.~Tian, M.~Lauer, and L.~Chen, ``Online multi-object tracking using joint
  domain information in traffic scenarios,'' \emph{IEEE Trans. Intell. Transp.
  Syst.}, vol.~21, no.~1, pp. 374--384, 2019.

\bibitem{Ren2015NIPS}
S.~Ren, K.~He, R.~Girshick, and J.~Sun, ``Faster r-cnn: Towards real-time
  object detection with region proposal networks,'' in \emph{Proc. Int. Conf.
  Neural Inform. Process. Syst. (NIPS)}, vol.~28, 2015.

\bibitem{bernardin2008evaluating}
K.~Bernardin and R.~Stiefelhagen, ``Evaluating multiple object tracking
  performance: the clear mot metrics,'' \emph{Eurasip J. Image Video Process.},
  vol. 2008, pp. 1--10, 2008.

\bibitem{li2009learning}
Y.~Li, C.~Huang, and R.~Nevatia, ``Learning to associate: Hybridboosted
  multi-target tracker for crowded scene,'' in \emph{Proc. IEEE Conf. Comput.
  Vis. Pattern Recognit. (CVPR)}, 2009, pp. 2953--2960.

\bibitem{luiten2021hota}
J.~Luiten, A.~Osep, P.~Dendorfer, P.~Torr, A.~Geiger, L.~Leal-Taix{\'e}, and
  B.~Leibe, ``Hota: A higher order metric for evaluating multi-object
  tracking,'' \emph{Int. J. Comput. Vis.}, vol. 129, no.~2, pp. 548--578, 2021.

\bibitem{lang2019pointpillars}
A.~H. Lang, S.~Vora, H.~Caesar, L.~Zhou, J.~Yang, and O.~Beijbom,
  ``Pointpillars: Fast encoders for object detection from point clouds,'' in
  \emph{Proc. IEEE Conf. Comput. Vis. Pattern Recognit. (CVPR)}, 2019, pp.
  12\,697--12\,705.

\bibitem{ren2017accurate}
J.~Ren, X.~Chen, J.~Liu, W.~Sun, J.~Pang, Q.~Yan, Y.-W. Tai, and L.~Xu,
  ``Accurate single stage detector using recurrent rolling convolution,'' in
  \emph{Proc. IEEE Conf. Comput. Vis. Pattern Recognit. (CVPR)}, 2017, pp.
  5420--5428.

\end{thebibliography}
	\vspace{-10mm}
\begin{IEEEbiography}[{\includegraphics[width=1in,height=1.25in,clip,keepaspectratio]{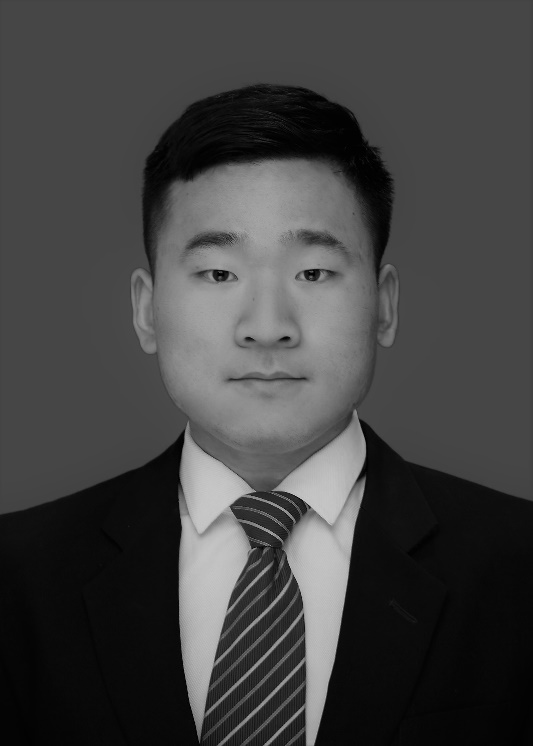}}]{Guangming Wang} (Graduate Student Member,
	IEEE) received the B.S. degree from Department of Automation from Central South University, Changsha, China, in 2018. He is currently pursuing the Ph.D. degree in Control Science and Engineering with Shanghai Jiao Tong University. His current research interests include SLAM and computer vision,  in particular, multi-object detection and tracking.
\end{IEEEbiography}

	\vspace{-10mm}
\begin{IEEEbiography}[{\includegraphics[width=1in,height=1.25in,clip,keepaspectratio]{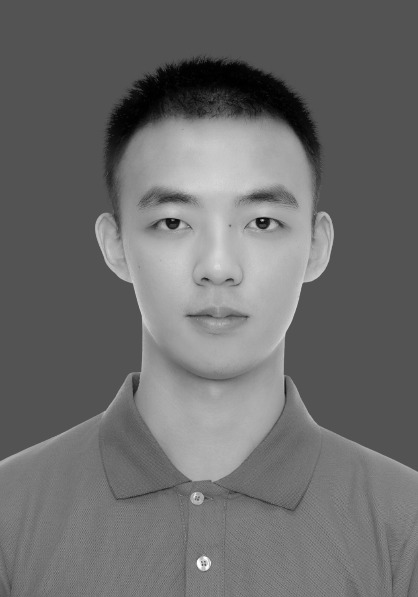}}]{Chensheng Peng} is currently pursuing the B.S. degree with the Department of Automation, Shanghai Jiao Tong University. His current research interests include SLAM and computer vision.
\end{IEEEbiography}
	\vspace{-10mm}
\begin{IEEEbiography}[{\includegraphics[width=1in,height=1.25in,clip,keepaspectratio]{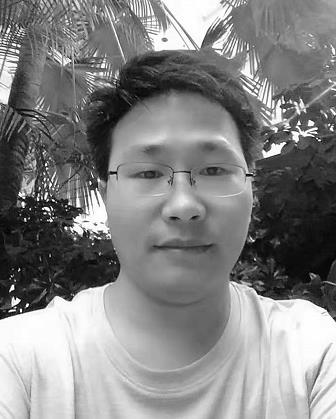}}]
	{Jinpeng Zhang} received B.S. and M.S. degrees in Material Science \& Engineering from Huazhong University of 
	Science \& Technology (HUST) in 2011 and 2014, respectively. From 2015–2019, he studied for his Ph.D. in 
	Pattern Recognition and Intelligent Systems at the National Laboratory of Pattern Recognition, Institute of 
	Automation, Chinese Academy of Sciences (NLPR, CASIA). His major research areas are machine learning and computer vision.
\end{IEEEbiography}
    \vspace{-10mm}
\begin{IEEEbiography}[{\includegraphics[width=1in,height=1.25in,clip,keepaspectratio]{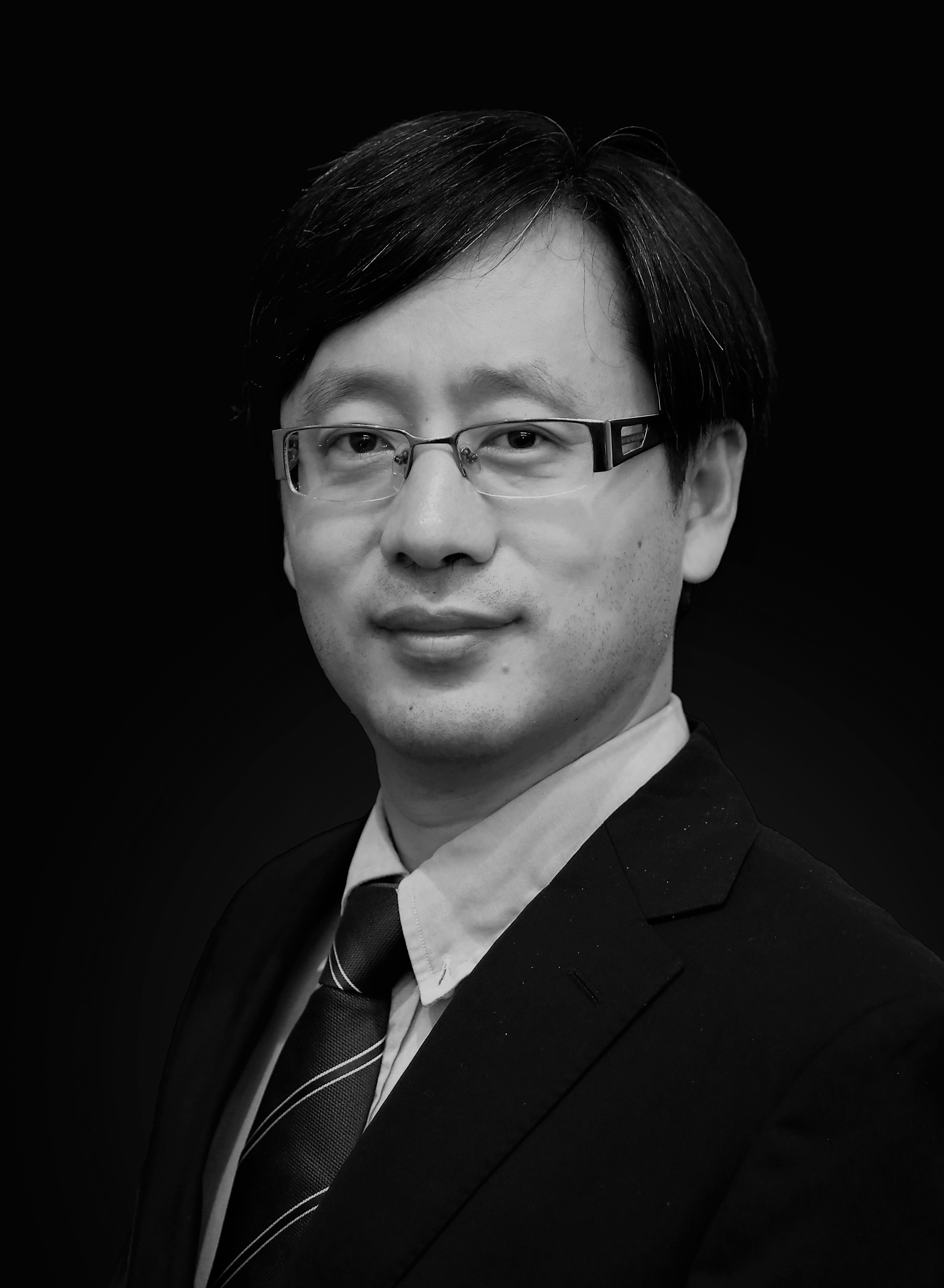}}]{Hesheng Wang}
 (Senior Member, IEEE) received the B.Eng. degree in electrical engineering from the Harbin Institute of Technology, Harbin, China, in 2002, and the M.Phil. and Ph.D. degrees in automation and computer-aided engineering from The Chinese University of Hong Kong, Hong Kong,
 in 2004 and 2007, respectively. He is currently a Professor with the Department of Automation, Shanghai
 Jiao Tong University, Shanghai, China. His current research interests include visual servoing, service robot, computer vision, and autonomous driving. He was the General Chair of the IEEE RCAR 2016, and the Program Chair of the IEEE ROBIO 2014 and IEEE/ASME AIM 2019. He has served as an Associate Editor for the IEEE TRANSACTIONS ON ROBOTICS from 2015 to 2019. He is an Associate Editor of IEEE TRANSACTIONS ON AUTOMATION SCIENCE AND ENGINEERING, IEEE ROBOTICS AND AUTOMATION LETTERS, Assembly Automation, and the International Journal of Humanoid Robotics; and a Technical Editor of the IEEE/ASME TRANSACTIONS ON MECHATRONICS. 
\end{IEEEbiography}

\end{document}